\documentclass[sigconf, nonacm]{acmart}

\usepackage{hyperref}
\usepackage{subfig}
\usepackage{graphicx}
\usepackage{float}
\usepackage{array}
\usepackage{algorithm}
\usepackage{algorithmic}
\usepackage{multirow}
\usepackage{amsmath,amsfonts,bm}

\def\eqref#1{equation~\ref{#1}}
\def\1{\bm{1}}

\DeclareMathAlphabet{\mathsfit}{\encodingdefault}{\sfdefault}{m}{sl}
\SetMathAlphabet{\mathsfit}{bold}{\encodingdefault}{\sfdefault}{bx}{n}

\AtBeginDocument{%
  }

\copyrightyear{2024}
\acmYear{2024}
\setcopyright{acmlicensed}
\acmConference[MMAsia '24]{Proceedings of the ACM Multimedia Asia 2024}{December 3-6, 2024}{Auckland, New Zealand}
\acmBooktitle{Proceedings of the ACM Multimedia Asia 2024 (MMAsia '24), December 3-6, 2024, 2024, Auckland, New Zealand}
\begin{document}

%%
%% The "title" command has an optional parameter,
%% allowing the author to define a "short title" to be used in page headers.
% \title{Multimodal Plant Disease Image Retrieval in the Wild}
\title{Snap and Diagnose: An Advanced Multimodal Retrieval System for Identifying Plant Diseases in the Wild}

%%
%% The "author" command and its associated commands are used to define
%% the authors and their affiliations.
%% Of note is the shared affiliation of the first two authors, and the
%% "authornote" and "authornotemark" commands
%% used to denote shared contribution to the research.

\author{Tianqi Wei, Zhi Chen, Xin Yu}
\affiliation{%
  \institution{The University of Queensland}
  \city{Brisbane}
  \country{Australia}}
\email{{tianqi.wei, zhi.chen, xin.yu}@uq.edu.au}

% \author{Tianqi Wei}
% \orcid{0009-0005-0134-6438}
% \affiliation{%
%   \institution{The University of Queensland}
%   \city{Brisbane}
%   \country{Australia}}
% \email{tianqi.wei@uq.edu.au}

% \author{Zhi Chen}
% \orcid{0000-0002-9385-144X}
% \affiliation{%
%  \institution{The University of Queensland}
%  \city{Brisbane}
%  \country{Australia}}
%  \email{zhi.chen@uq.edu.au}

% \author{Xin Yu}
% \orcid{0000-0002-0269-5649}
% \affiliation{%
%   \institution{The University of Queensland}
%   \city{Brisbane}
%   \country{Australia}}
%   \email{xin.yu@uq.edu.au}
% \author{Zi Huang}
% \orcid{0000-0002-9738-4949}
% \affiliation{%
%  \institution{The University of Queensland}
%  \city{Brisbane}
%  \country{Australia}}
%  \email{helen.huang@uq.edu.au}

%%
%% By default, the full list of authors will be used in the page
%% headers. Often, this list is too long, and will overlap
%% other information printed in the page headers. This command allows
%% the author to define a more concise list
%% of authors' names for this purpose.
% \renewcommand{\shortauthors}{Trovato et al.}

%%
%% The abstract is a short summary of the work to be presented in the
%% article.
\begin{abstract}
% Plant disease recognition is a critical task that ensures crop health and mitigates the damage caused by diseases. Enabling farmers to .... 
% [Farmers usually use the observed symptoms to obtain an existing image with either images or textual descriptions such as ``yellow spots on leaves" or ``wilting flowers". Existing systems either only allow content-based retrieval or cannot handle in-the-wild images.] In this paper, we develop a multimodal plant disease image retrieval system to support image search based on either image or text prompts. Specifically, we utilize the largest in-the-wild plant disease dataset PlantWild, which includes over 18,000 images across 88 categories, to provide diverse retrieval results. Then, to achieve cross-modal retrieval, we utilize a CLIP-based vision-language model that encodes both disease descriptions and disease images into the same latent space. Built on top of the retriever, our web application allows users to upload either plant disease images or disease descriptions to retrieve the corresponding images with similar characteristics.

Plant disease recognition is a critical task that ensures crop health and mitigates the damage caused by diseases. A handy tool that enables farmers to receive a diagnosis based on query pictures or the text description of suspicious plants is in high demand for initiating treatment before potential diseases spread further. In this paper, we develop a multimodal plant disease image retrieval system to support disease search based on either image or text prompts. Specifically, we utilize the largest in-the-wild plant disease dataset PlantWild, which includes over 18,000 images across 89 categories, to provide a comprehensive view of potential diseases relating to the query. Furthermore, cross-modal retrieval is achieved in the developed system, facilitated by a novel CLIP-based vision-language model that encodes both disease descriptions and disease images into the same latent space. Built on top of the retriever, our retrieval system allows users to upload either plant disease images or disease descriptions to retrieve the corresponding images with similar characteristics from the disease dataset to suggest candidate diseases for end users' consideration.

\end{abstract}

%%
%% The code below is generated by the tool at http://dl.acm.org/ccs.cfm.
%% Please copy and paste the code instead of the example below.
%%

%%
%% The code below is generated by the tool at http://dl.acm.org/ccs.cfm.
%% Please copy and paste the code instead of the example below.
%%


%%
%% Keywords. The author(s) should pick words that accurately describe
%% the work being presented. Separate the keywords with commas.
%% A "teaser" image appears between the author and affiliation
%% information and the body of the document, and typically spans the
%% page.

% \received{20 February 2007}
% \received[revised]{12 March 2009}
% \received[accepted]{5 June 2009}

%%
%% This command processes the author and affiliation and title
%% information and builds the first part of the formatted document.
\maketitle

\section{Introduction}
With the global population on the rise, the demand for food continues to escalate \cite{oerke2012cropdemand}. Plant diseases significantly reduce crop yield reduction, inflicting economic losses exceeding \$200 billion annually \cite{plant_pathology}. Plant disease recognition plays a vital role in crop protection against diseases, as it directly impacts agricultural sustainability and global food security. Traditionally, plant disease recognition relies on experienced farmers or agricultural experts for manual identification, but these practices are time-consuming, costly, and not always available. 
In this context, automatic disease recognition with machine learning approaches has drawn much attention in the plant pathology community. 
While existing methods have achieved promising results on in-laboratory images \cite{plantvillage,cassava,citrus_detection}, their performance significantly declines when applied to images captured in the wild.
% due to the domain gap, as plant diseases manifest differently across various regions and crops, adding to the challenge of disease recognition.
Furthermore, farmers often need to match observed symptoms, such as ``yellow spots on leaves" or ``wilting flowers", with corresponding images. It is thus desirable to achieve cross-modal disease retrieval with textual queries.

% [From manual to automatic] Intuitively, automatic ... image ... . So content-based .... 
% [Automatic to in-the-wild]
% [Talk about multimodal(content-based to multimodal)] Farmers typically need to match observed symptoms, such as ``yellow spots on leaves" or ``wilting flowers", with corresponding images.

\begin{figure}[tbp]
    \centering
    \includegraphics[width=\linewidth]{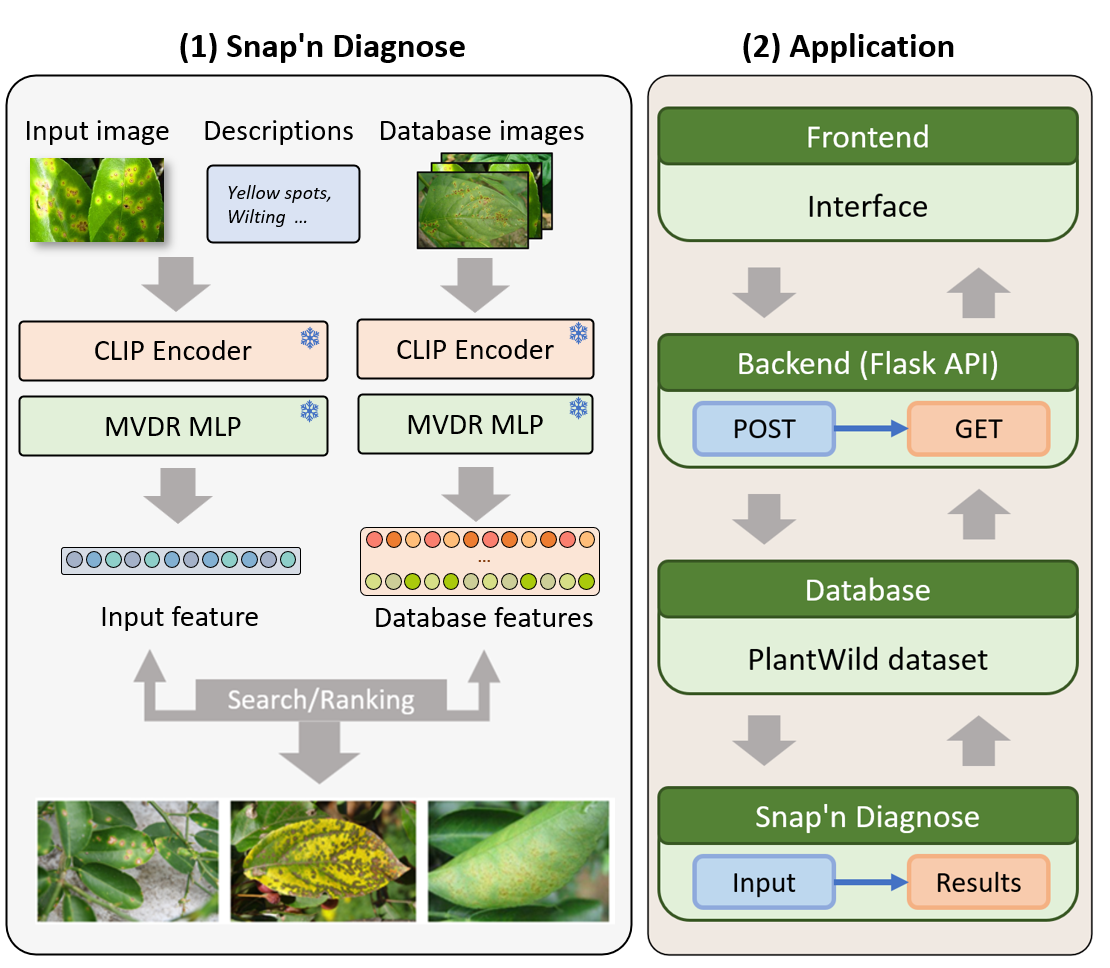}
    \vspace{-10pt}
    \caption{Overview of Snap'n Diagnose. We leverages CLIP \cite{CLIP} and MVPDR \cite{MVPDR} to extract visual/text features from images and texts and conduct multimodal image retrieval for identifying plant disease in the wild.}
    \label{fig:framework}
    \vspace{-5pt}
\end{figure}

To facilitate this need, current plant disease image retrieval systems predominantly support unimodal (image-only) queries \cite{singh2020review,baquero2014image} and are limited by plant types \cite{baquero2014image,zhijun2015research} and inability to handle in-the-wild images \cite{peng2022leaf,plantvillage,cassava,citrus_detection}. These constraints hinder their practical utility in diverse agricultural settings.

To address these limitations, in this paper, we propose a multimodal image retrieval system for plant diseases in the wild. Specifically, to accommodate the needs for diverse plant diseases, we construct our retrieval database with the world-largest plant disease dataset PlantWild \cite{MVPDR}. It includes over 18,000 in-the-wild plant images across 89 classes. Notably, PlantWild dataset provides diverse textual descriptions for each disease type. Further, to facilitate cross-modal queries, we develop a CLIP-based \cite{CLIP} retrieval method that projects both images and textual descriptions of each disease into the same latent space. Such a shared latent space allows users to retrieve the image samples closest to the query. Built upon this retrieval model, we then develop a retrieval system \textbf{Snap'n Diagnose} to provide an interface for retrieval interaction.

% The use procedure of Snap'n Diagnose involves a user to type in a textual query or upload a query picture that potentially has diseases present. The system will then project the uploaded information into query representations to compute the similarity with candidate images in the database. The most similar results will be presented to the user.

Snap'n Diagnose offers a user-friendly interface that simplifies the retrieval process. Users can either type in a textual description of observed symptoms or upload a photo of the affected plant. The system then transforms this input into query representations and calculates similarities with images in the database. The results, ranked by relevance, are promptly displayed, providing users with reliable and actionable insights into potential plant diseases.

% Snap'n Diagnose allows users to upload an image of a plant disease, which is then processed to extract feature vectors. These vectors are compared against the database using cosine similarity to identify and rank the most relevant images. This method ensures users receive the most accurate matches to their queries, thereby enhancing decision-making in crop management.

% \vspace{-8pt}
\section{System}
\subsection{Methodology}
The retrieval model in Snap'n Diagnose is based on the MVPDR method \cite{MVPDR}, which is designed for in-the-wild plant disease recognition and has proved effective in adapting images across different environments. Therefore, it is suitable for searching plant disease images. Besides, it also benefits from the image search engine \cite{cvpr20_tutorial_image_retrieval}.

\begin{figure}[tbp]
    \centering
    \includegraphics[width=\linewidth]{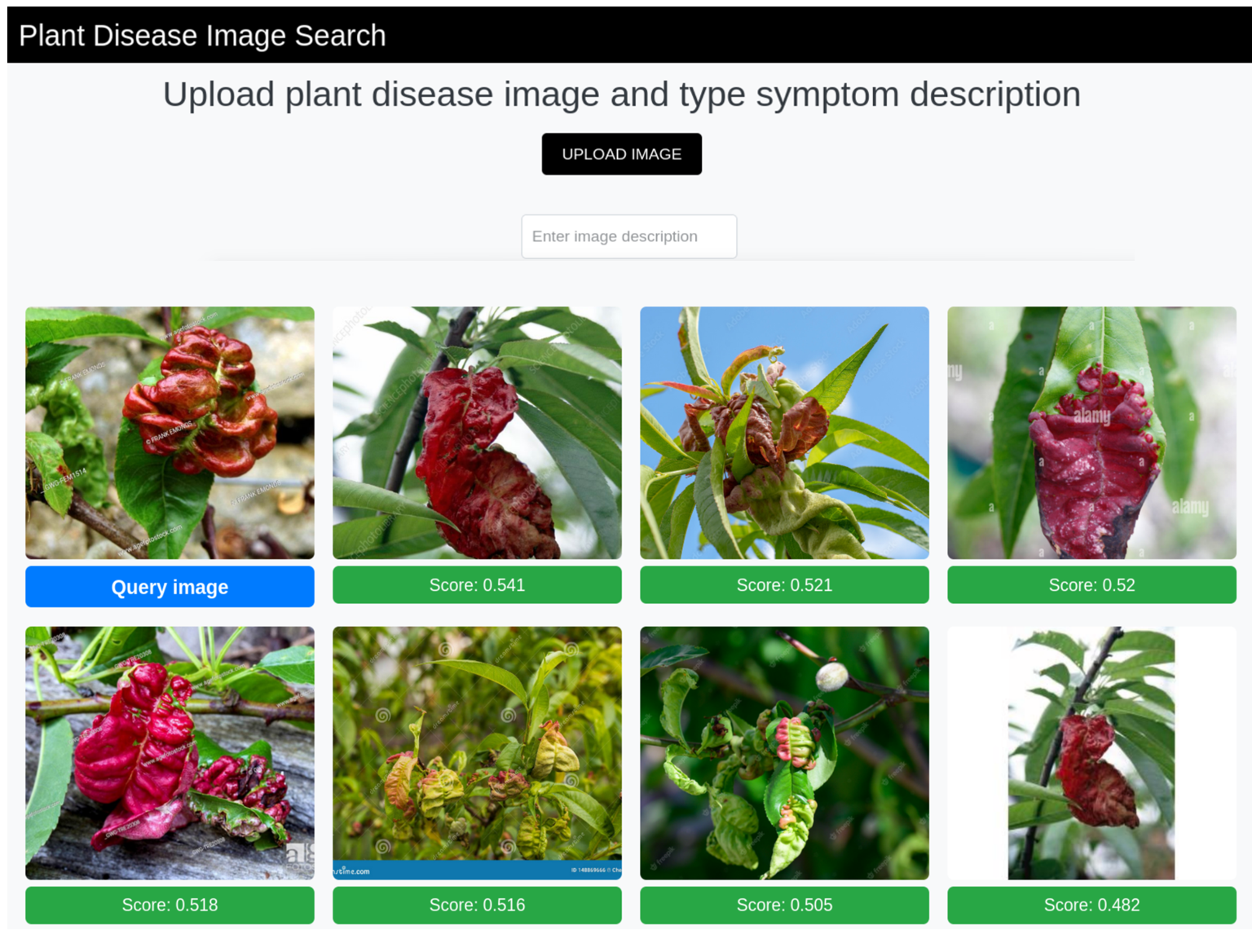}
    \caption{A screenshot of our system interface. After receiving the image/textual queries, Snap'n Diagnose results will be returned and ranked in order of cosine similarity.}
    % \vspace{-5pt}
    \label{fig:interface}
    % \vspace{-5pt}
\end{figure}

Snap'n Diagnose extracts and stores visual/text features with CLIP encoders and MVPDR's fine-tuned MLP. These obtained image features preserve a wide range of plant disease features due to the pre-trained CLIP and the fine-tuned MLP. During the inference process, a given input image or textual query will first be extracted into a feature vector. Afterward, the cosine similarity between the vector and all the stored features can be acquired. The outcome is a ranked list that reflects the most pertinent results corresponding to the input image. The workflow is presented in Figure \ref{fig:framework}.

\vspace{-5pt}
\subsection{Implementation}
The retrieval system has been designed for users to interactively conduct multimodal image retrieval, as presented in Figure \ref{fig:interface}. The system consists of a backend component to perform multimodal retrieval and a frontend interface for user interaction.

\vspace{5pt}
\noindent \textbf{Backend.} \hspace{5pt}
% \subsubsection{\textbf{Backend}}
We deploy Snap'n Diagnose with a Flask API in the backend. The API receives the uploaded images/texts as queries and performs inference, searching for images with similar symptoms in a database for responses.
We leverage PlantWild \cite{MVPDR} as our database, which is the largest plant disease image dataset that includes a vast collection of images captured in diverse environments. Specifically, we extract visual features using CLIP's image encoder and MVPDR's pre-trained MLP, utilizing these features for inference. This operation not only reduces CPU usage but also removes the time for models to process images. Therefore, our system has a fast response time, providing better user experiences.

\begin{table}[tbp]
    \centering
        \caption{Performance comparison of multimodal image retrieval for plant diseases between different models.}
    \begin{tabular}{cccccc}
    \toprule
         Methods & Top-1 & Top-5 & Top-10 & mAP\\
         \midrule
         Zero-shot CLIP \cite{CLIP} & 40.92 & 65.75 & 74.81 & 68.72\\
         Snap'n Diagnose (ours) & \textbf{67.32} & \textbf{80.65} & \textbf{88.11} & \textbf{79.34}\\
         \bottomrule
    \end{tabular}
    % \vspace{5pt}

    \label{tab:retrieval performance}
    \vspace{-0.5em}
\end{table}

\vspace{5pt}
% \subsubsection{\textbf{Frontend interface}}
\noindent \textbf{Frontend interface.} \hspace{5pt}
The interface of our system is presented in Figure \ref{fig:interface}. It is accessible through web browsers on both PC and mobile devices. The design is straightforward, allowing users to easily use it without extensive knowledge. When encountering unknown plant diseases in the wild, users can simply take a photo or typing the symptom descriptions and upload it via the interface. After receiving the request, the backend will perform cross-modal retrieval to obtain similar image results and then return them to the frontend interface. The result images will be arranged following the query image in descending order of their similarity scores, and the corresponding scores will be displayed below each image.

% \vspace{-5pt}
\vspace{-3pt}
\section{Experiment}
We evaluate the performance of the Snap'n Diagnose in plant disease image retrieval and compare its performance with the pre-trained CLIP vision-language model. 
Experiments are conducted on the PlantWild dataset.
According to the results presented in Table \ref{tab:retrieval performance}, our method exhibits excellent performance and consistently outperforms Zero-shot CLIP across all evaluation metrics, including Top-1, Top-5, Top-10 accuracy, and mean Average Precision (mAP). These results underscore the effectiveness of Snap'n Diagnose in retrieving relevant plant disease images, demonstrating that it can offer a practical tool in plant disease recognition.

\section{Conclusion}
In this paper, we present Snap'n Diagnose, a multimodal image retrieval system designed for identifying plant disease in-the-wild. 
It addresses the limitations of existing retrieval systems that only support unimodal, laboratory and single plant types. 
Further, Snap'n Diagnose enables farmers to receive plant diagnosis based on query pictures or the textual description of suspicious symptoms. 

%%%%%%%%%%%%%%%%%%%%%%%%%%%%%%%%%%%%%%%%%%%%%%%%%%%%%%%%%%%%%%%%%%%%%%%%%%%%%%%%%%%%%%%%%%%%%%%%%%%%%%%

%%
%% The next two lines define the bibliography style to be used, and
%% the bibliography file.
% \newpage
\bibliographystyle{ACM-Reference-Format}
\balance
\bibliography{main}

%%
%% If your work has an appendix, this is the place to put it.
% \appendix

% \section{Research Methods}

% \subsection{Part One}

% Lorem ipsum dolor sit amet, consectetur adipiscing elit. Morbi
% malesuada, quam in pulvinar varius, metus nunc fermentum urna, id
% sollicitudin purus odio sit amet enim. Aliquam ullamcorper eu ipsum
% vel mollis. Curabitur quis dictum nisl. Phasellus vel semper risus, et
% lacinia dolor. Integer ultricies commodo sem nec semper.

% \subsection{Part Two}

% Etiam commodo feugiat nisl pulvinar pellentesque. Etiam auctor sodales
% ligula, non varius nibh pulvinar semper. Suspendisse nec lectus non
% ipsum convallis congue hendrerit vitae sapien. Donec at laoreet
% eros. Vivamus non purus placerat, scelerisque diam eu, cursus
% ante. Etiam aliquam tortor auctor efficitur mattis.

% \section{Online Resources}

% Nam id fermentum dui. Suspendisse sagittis tortor a nulla mollis, in
% pulvinar ex pretium. Sed interdum orci quis metus euismod, et sagittis
% enim maximus. Vestibulum gravida massa ut felis suscipit
% congue. Quisque mattis elit a risus ultrices commodo venenatis eget
% dui. Etiam sagittis eleifend elementum.

% Nam interdum magna at lectus dignissim, ac dignissim lorem
% rhoncus. Maecenas eu arcu ac neque placerat aliquam. Nunc pulvinar
% massa et mattis lacinia.

\end{document}